# POLARITY DETECTION OF MOVIE REVIEWS IN HINDI LANGUAGE

Richa Sharma[1], Shweta Nigam[2] and Rekha Jain[3]

[1,2]M.Tech Scholar, Banasthali Vidyapith, Rajasthan, India.
[3]Assistant Professor, Banasthali Vidyapith, Rajasthan, India.

## ABSTRACT

*Nowadays peoples are actively involved in giving comments and reviews on social networking websites and other websites like shopping websites, news websites etc. large number of people everyday share their opinion on the web, results is a large number of user data is collected .users also find it trivial task to read all the reviews and then reached into the decision. It would be better if these reviews are classified into some category so that the user finds it easier to read. Opinion Mining or Sentiment Analysis is a natural language processing task that mines information from various text forms such as reviews, news, and blogs and classify them on the basis of their polarity as positive, negative or neutral. But, from the last few years, user content in Hindi language is also increasing at a rapid rate on the Web. So it is very important to perform opinion mining in Hindi language as well. In this paper a Hindi language opinion mining system is proposed. The system classifies the reviews as positive, negative and neutral for Hindi language. Negation is also handled in the proposed system. Experimental results using reviews of movies show the effectiveness of the system.*

## 1.INTRODUCTION

Online shopping is very common now, peoples find online shopping very easy task, they can buy anything just sitting in a home with the help of simple click. Buyers allow their customers to share their opinions in the forms of reviews, so that with the help of these reviews they can know about the likes and dislikes of their products. All this is possible with the help of the Internet, but user reviews are increasing at a faster rate, everyday large number of customers write their opinions about the products on these websites, which makes it difficult for the user to read all the reviews and would take the decision. It is important to mine these reviews and identify their opinions expressed in these reviews. Opinion Mining is a *Natural Language Processing (NLP)* and *Information Extraction (IE)* task that aims to obtain feelings of the writer expressed in positive or negative comments by analyzing various text forms such as reviews, news, and blogs [10]. Opinion Mining can be defined as a sub-discipline of computational linguistics that focuses on extracting opinion of persons from the web. It combines the techniques of computational linguistics and Information Retrieval (IR).Opinion Mining is performed at one of the three levels:

- **Document Level** determines the polarity of whole document e.g. the document is given as

     

न ही फिल्म का कॉन्सेप्ट नया है और न ही फिल्म में ज्यादा मजेदार कॉमेडी है। फिल्म में आयुष्मान और सोनम की खराब केमिस्ट्री देखने को मिली है। हां, आयुष्मान की ऋषि के साथ जबरदस्त केमिस्ट्री दिखी।फिल्म में कई खामियां हैं.

At document level this document is classified in the category of negative opinion.

- **Sentence level** determines the polarity of sentences eg the sentences given below are classified as

    1. यह मोबाइल फोन अच्छा है | This sentence is classified as positive
    2. इस होटल का खाना खराब है | This sentence is classified as negative
    3. में सुबह घूमने जाता हूँ | This sentence is classified as neutral

- **Aspect level** determines the polarity of sentences/documents for each feature it contains. Eg the sentences given below are classified at aspect level as

    Feature: camera

    1. Positive sentences
        I. फोन में 24 मेगापिक्सेल का कैमरा है जो लाजवाब है |
    2. Negative Sentences
        I. रात में इस फोन के कैमरा से धुंदले फोटोस आते है|

Mostly research work in Opinion Mining is carried out in English language. But from the last few years, Hindi content has also been available on the web and increasing at a faster rate. There is a need to perform opinion mining in Hindi language so that the customer reviews in Hindi can be easily classified and proved useful for the users in decision making. But performing opinion mining in Hindi language is not an easy task, there are lots of challenges comes in the way to perform this task which are followed:

- Sufficient resources for Hindi language are not available. Annotated corpora and tagger for Hindi language is not as good compared to English language makes the sentiment analysis task time consuming.
- Hindi is a free word order language means there is no specific arrangement of words in Hindi language i.e. subject, object and verb comes in any order whereas English is fixed word order language i.e. subject is always followed by a verb and then followed by an object. Word order is important for determining the polarity of given text.
- Same words in Hindi language having same meaning may occur in multiple contexts, it is impossible that the lexicon contains all the possible words.

In this paper a Hindi language based Opinion Mining System is proposed named as "Hindi Sentiment Orientation System" based on an unsupervised dictionary approach that determine the polarity of user reviews in Hindi language. The Hindi dictionary has developed by us that contain the most frequently used Hindi words and its synonyms and antonyms. Proposed methodology also handles negation. The appropriate polarity of the reviews is given based on negation. The rest of the paper is organized as follows: Section 2 describes the related work performed in Hindi language. Section 3 explains the proposed approach. Experimental Results are presented in Section 4. The last concludes the study.

## 2. EXISTING RESEARCH WORK

Small amount of work has been done in opinion mining for Hindi language which are as follows: Researches were carried out in Hindi and Bengali language. The most prominent work has been done by Amitava Das and Bandopadhya [1],they developed sentiwordnet for Bengali language.To obtain a Bengali SentiWordNet, Word level lexical-transfer technique has been applied to each entry in English SentiWordNet using an English-Bengali Dictionary. 35,805 Bengali entries have been returned by their experiment.

To predict the sentiment of a word four strategies were devised by Das and Bandopadhya [2]. An interactive game was proposed by them in the first approach in which words were annotated along with their polarity. In Second approach, to determine the polarity of a word Bi-Lingual dictionary for English and Indian Languages were used. Wordnet was used in third approach and by using synonym and antonym relations polarity was determined. To determine the polarity of words learning from pre-annotated corpora takes place in Fourth approach;.

Dipankar Das and Bandopadhya [11], identified emotional expressions in Bengali corpus. Emotional components such as holders, intensity and topics were taken to identify the emotional expression. They classified the words in six emotion classes and with three types of intensities to perform sentence level annotation.

Fallback strategy was proposed by Joshi et al. [3] for Hindi language. By using three approaches: In-language Sentiment Analysis, Machine Translation and Resource Based Sentiment Analysis in this strategy, a lexical resource were developed by them in, Hindi SentiWordNet (HSWN) based on its English format. H-SWN (Hindi-SentiWordNet) was created by them by using two lexical resources (English SentiWordNet and English-Hindi WordNet Linking [20]). Words in English SentiWordNet were replaced by equivalent Hindi words to get H-SWN by using Wordnet linking. 78.14 accuracy was achieved by their experiment.

The lexicon was created by Bakliwal et al.[4] using a graph based method .They determine that how the synonym and antonym relations can be used to generate the subjectivity lexicon by using the simple graph traversal approach. 79% accuracy was achieved on classification of reviews by their proposed algorithm. Mukherjee et al. [26] showed that by incorporating discourse markers in a bag-of-words model improves the sentiment classification accuracy by 2 - 4%. Bakliwal et al. [5] devised a new scoring function to classify Hindi reviews as positive or negative and test on two different approaches. Combination of simple N-gram and POS Tagged N-gram approaches were also used by them.

A novel approach was proposed by Ambati et al. [7] to detect errors in the treebanks. Validation time was significantly reduced by this approach. This approach detects 76.63% of errors at the dependency level when tested on Hindi dependency Treebank. A Graph based method was proposed by Piyush Arora et al. [25] to build a subjective lexicon for Hindi language, using WordNet as a resource. Small seed list of opinion words was initially built and by using WordNet, synonyms and antonyms of the opinion words were determined and added to the seedlist. Wordnet was traversed like a graph where every word was considered as a node, which is connected to their synonyms and antonyms.74% accuracy was achieved by their experiment on classification of reviews

An efficient approach was developed by Namita mittal et al.[22] based on negation and discourse relation to identifying the sentiments from Hindi content. The annotated corpus for Hindi language was developed and existing Hindi SentiWordNet (HSWN) was improved by incorporating more opinion words into it. They also devised the rules for handling negation and discourse that affect the sentiments expressed in the review. 80% accuracy was achieved by their proposed algorithm for classification of reviews.

## 3. PROPOSED WORK

The proposed approach for opinion mining in Hindi language is closely related to the Minqing Hu and Bing Liu work on Mining and Summarizing Customer Reviews [11]. But instead of using the Wordnet, Hindi dictionary was developed by us to determine the polarity of Hindi reviews. Figure 1 gives the overview of the proposed system. User and critic reviews of the movies were collected and applied as an input to the system. The system classifies each review as positive, negative and neutral and presents the total number of positive, negative and neutral number of sentences separately in the output. The output generated by the system is helpful for the users in decision making; they can easily identify how many positive and negative sentences are present. The polarity of the given sentences is determined on the basis of the majority of opinion words.

The system is divided into following phases.

### 1. Data Collection for Hindi language

To perform opinion mining in Hindi language, the data set has to be prepared first. To prepare the data set, large numbers of Hindi reviews were collected from the Web. There are lots of websites like which contain Hindi content. Here, Movie reviews were collected from the Hindi newspapers website. But before applying as an input, the collected data first preprocessed. After preprocessing the reviews were applied as an input.

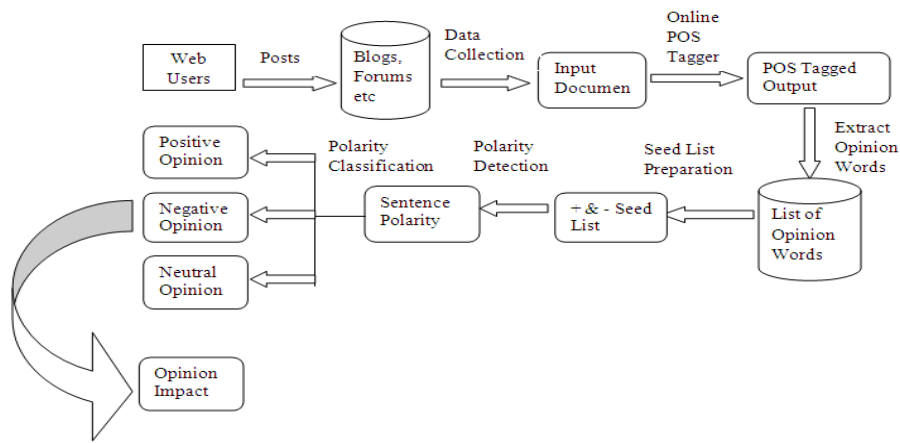

Figure 2 Hindi Sentiment Orientation System.

## 1. Part of Speech Tagging

POS tagging is very important for opinion mining. POS tagging is used to determine the opinion words and features in the reviews. POS tagging can be done manually or with the help of POS tagger. POS tagger tag all the words of reviews to their appropriate part of speech tag. Manual POS tagging of the reviews takes lots of time. Here, Online POS tagger of Hindi is used to tag all the words of reviews. e.g.

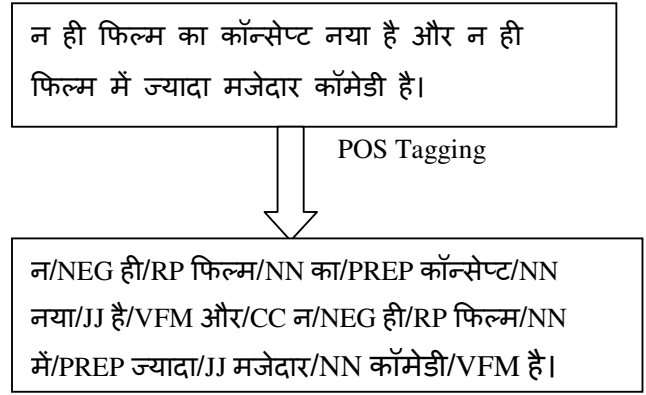

Figure 2 Example of POS Tagging

## 1. Opinion words extraction and Seed list preparation

Seed list is prepared first in which most frequently used Hindi words along with their polarity are stored. All the opinion words which were extracted after the POS tagging are first matched with the stored words in the seed list if it is matched with the words stored in the seed list then there is no need to determine the synonyms of the word. But if the word is not found in the seed list then the synonyms of that word are determined with the help of Hindi dictionary that is also built by us. Each synonym is matched with the words in the seed list, if any synonym is matched the opinion word along with its synonyms is stored in the seed list with same polarity. It grows every time whenever synonyms words found in Hindi dictionary are matched with seed list.

## 1. Polarity detection of reviews

In the last phase, the polarity of the collected reviews is determined with the help of seed list and Hindi dictionary. The polarity of the reviews is determined on the basis of majority of opinion words, if positive words are more in the review than the polarity of the review is positive otherwise it is negative. If positive and negative words are equal in a review the polarity is neutral. As negation is also handled in this approach, so if the opinion word is followed by not then the polarity of review is reversed. e.g. the sentence.

अभिजीत यह गाना अच्छा नहीं गा पाये | Here, the opinion word is 'अच्छा' which is followed by 'नहीं' shows negative polarity. Figure3 gives an example of how the proposed system classifies the Hindi reviews.

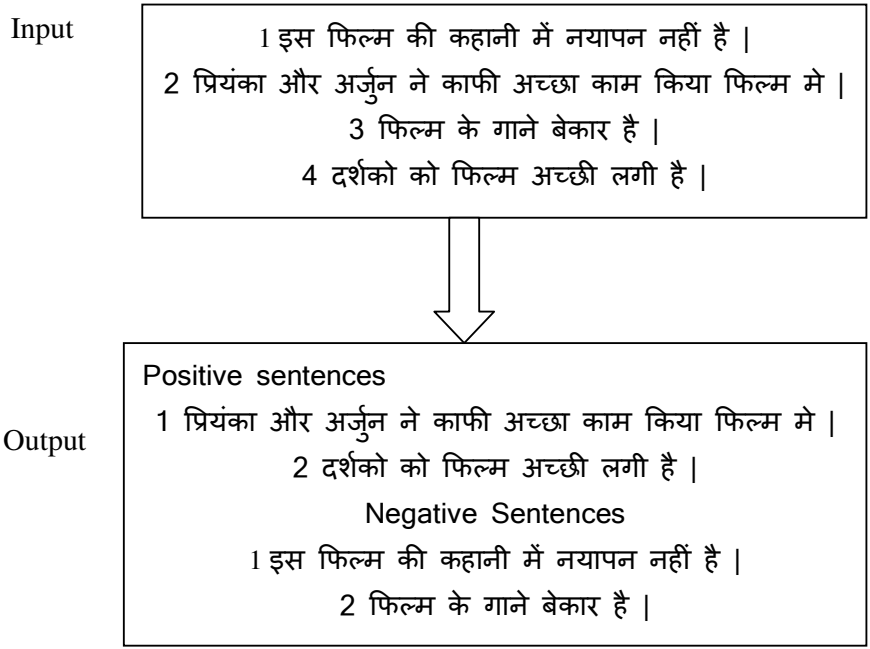

Input

Output

Figure 3 Example of Opinion Mining

## 4. EXPERIMENTS & RESULTS

Experiment is conducted on movie reviews. Movie reviews were collected from several websites contain Hindi reviews. Reviews were applied as input to the system which classifies these reviews and determine the polarity of these reviews and present the summarized positive and negative results which prove to be helpful for the users. Input reviews were also classified by us to determine how well the system classified the reviews as compared to human judgement. Three evaluation measures are used on the basis of which system performance is computed, these are:

- Precision
- Recall
- Accuracy

The common way for computing these measures is based on the confusion matrix shown in Table 1.

| Instances | Predicted positives | Predicted negatives |
|---|---|---|
| **Actual positive instances** | # of True positive instances (TP) | # of false negative instances (FN) |
| **Actual negative instances** | # of false positive instances (FP) | # of True Negative instances (TN) |

Table 1. Confusion Matrix

- *Accuracy* is the portion of all true predicted instances against all predicted instances. An accuracy of 100% means that the predicted instances are exactly the same as the actual instances.

$$\text{Accuracy} = \frac{tp+tn}{tp+tn+fp+fn} \qquad \text{eq (1)}$$

- *Precision* is the portion of true positive predicted instances against all positive predicted instances.

$$\text{Precision} = \frac{tp}{tp+fp} \qquad \text{eq (2)}$$

- *Recall* is the portion of true positive predicted instances against all actual positive instances.

$$\text{Recall} = \frac{tp}{tp+fn} \qquad \text{eq (3)}$$

On the basis of these evaluation measures, results show that 'Hindi Sentiment Orientation System' is performed well in the movie review domain. The experiments have been performed by using 50 sentences of movie reviews.

Table 2 presents the precision, accuracy and recall results of current system
Figure 4 presents the precision, accuracy and recall results of current system in graphical form.

| Measures | Results |
|---|---|
| Accuracy | 0.65 |
| Precision | 0.66 |
| Recall | 0.78 |

Table 2 Hindi Sentiment Orientation System Results

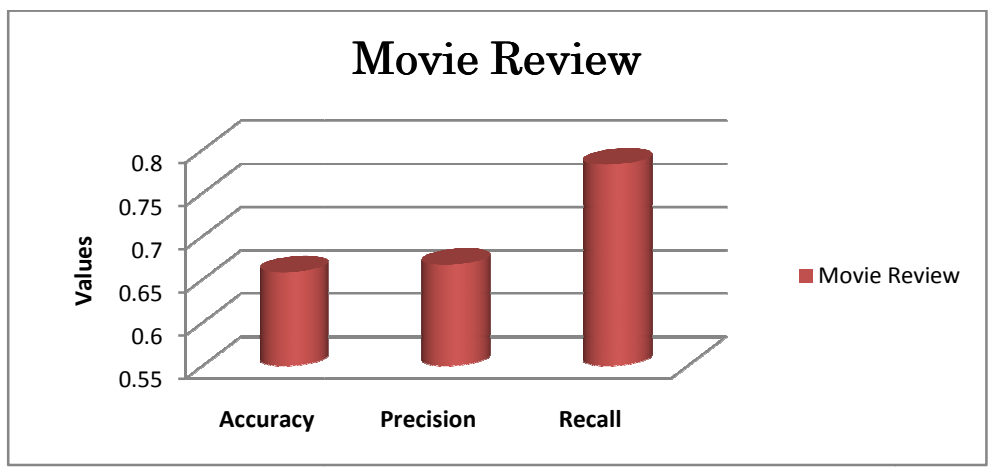

Figure 4 : Performance of Hindi Sentiment Orientation System

The above results show that the 'Hindi Sentiment Orientation System' performs well with respect to the movie review domain which proves that system is efficient. 'Hindi Sentiment orientation system' shows the accuracy of 65% which proves the system more efficient.

## 5. CONCLUSION

Opinion Mining is an emerging research field and this task is very important because peoples spent their most of the time on the web. In this paper an approach is proposed to determine the sentiment orientation i.e. polarity of the Hindi reviews. Opinion mining is needed to be performed in Hindi language because of the increase in Hindi data on the web. Separate positive and negative summarized results are generated which is helpful for the user in decision making. Experimental results indicate that the proposed approach is performing well in this domain and achieved the accuracy of 65%. In future the work can be extended to perform feature base opinion mining in Hindi reviews by extracting the feature from these reviews, and to perform opinion mining in other Hindi domain. Efforts would be done to improve the accuracy of the system by handling relative clauses like "सिर्फ – बल्कि" For example "फिल्म सिर्फ अच्छी ही नहीं बल्कि बेमिसाल है".